%% file: main_emecom2020.tex
\documentclass{article}

\PassOptionsToPackage{square,numbers}{natbib}

\usepackage[final]{neurips_2020_emecom}

\usepackage[utf8]{inputenc} 
\usepackage[T1]{fontenc}    
\usepackage{hyperref}       
\usepackage{url}            
\usepackage{booktabs}       
\usepackage{amsfonts}       
\usepackage{nicefrac}       
\usepackage{microtype}      

\usepackage[pdftex]{graphicx}
\usepackage{caption}
\usepackage{float}
\usepackage{subcaption}

\title{ReferentialGym: A Nomenclature and Framework for Language Emergence \& Grounding in (Visual) Referential Games}

\author{%
  Kevin Denamganaï and James Alfred Walker \\
  Department of Computer Science\\
  University of York\\
  York, UK \\
  \texttt{kyd500@york.ac.uk}, \texttt{james.walker@york.ac.uk} \\
  
}

\begin{document}

\maketitle

\input{parts/abstract}

\input{parts/introduction}

\input{parts/referential-games}

\input{parts/referential-gym}

\input{parts/related-future-works}

\input{parts/conclusion}

\section*{Broader Impact}

This work consists solely of simulations, thus evacuating some of the ethical concerns, as well as the concerns with regards to the consequences of failure of the system presented. With regards to the ethical aspects related to its inclusion in the field of Artificial Intelligence, we argue that our work aims to have positive outcomes on the development of human-machine interfaces, albeit being not yet mature enough to aim for this goal. The current state of our work does not allow us to extrapolate towards negative outcomes.

This work should benefit the research community of language emergence and grounding, in its current state.

\begin{ack}
This work was supported by the EPSRC Centre for Doctoral Training in Intelligent Games \& Games Intelligence (IGGI) [EP/L015846/1]. 

We would like to thank the anonymous reviewers for their constructive feedback on the draft of this paper.

We gratefully acknowledge the use of Python\cite{python-2009}, IPython\cite{ipython-perez-2007}, SciPy\cite{SciPy-NMeth2020}, Scikit-learn\cite{Scikit-learn:JMLR:v12:pedregosa11a}, Scikit-image\cite{scikit-image-van2014}, NumPy\cite{NumPy-Array2020}, Pandas\cite{pandas1-mckinney-proc-scipy-2010,pandas2-reback2020}, OpenCV\cite{opencv_library}, PyTorch\cite{pytorch-paszke-NEURIPS2019_9015}, TensorboardX\cite{huang2018tensorboardx}, and Tensorboard from the Tensorflow ecosystem\cite{tensorflow2015-whitepaper}, without which this work would not be possible.

\end{ack}

\bibliography{bibli}

\appendix

\end{document}

%% file: parts/abstract.tex
\begin{abstract}

Natural languages are powerful tools wielded by human beings to communicate information and co-operate towards common goals. Their values lie in some main properties like compositionality, hierarchy and recurrent syntax, which computational linguists have been researching the emergence of in artificial languages induced by language games. Only relatively recently, the AI community has started to investigate language emergence and grounding working towards better human-machine interfaces. For instance, interactive/conversational AI assistants that are able to relate their vision to the ongoing conversation. 

This paper provides two contributions to this research field. Firstly, a nomenclature is proposed to understand the main initiatives in studying language emergence and grounding, accounting for the variations in assumptions and constraints. Secondly, a PyTorch based deep learning framework is introduced, entitled ReferentialGym, which is dedicated to furthering the exploration of language emergence and grounding. By providing baseline implementations of major algorithms and metrics, in addition to many different features and approaches, ReferentialGym attempts to ease the entry barrier to the field and provide the community with common implementations.

\end{abstract}

%% file: parts/introduction.tex
\section{Introduction}

Natural languages, whose evolution is supported by a culture of individuals that speak them, are a cornerstone of our ability to communicate and co-operate among each other, with different levels of expressivity and/or conciseness. They also act as a media through which we build our own learned representation of the world, as we experience it through our other modalities, e.g. vision. 
Although very little is known about how they came to be such a useful media, it is recognised that their values lie in some of their main properties: compositionality, hierarchical and/or recurrent syntax. They contribute to our ability to express an infinity of meanings while only using a finite amount of symbols, i.e. words and letters. Computational linguists have been researching the emergence of these properties in artificial languages induced by language games~\citep{chomsky1976reflections, Kirby199812, brighton2001survival, Brighton&Kirby2006, Kirby&Hurford2002, hurford1987language, Smith2003, Kirby2014} to better understand the evolution of natural languages. It is only relatively recently that it has also been investigated within the context of deep learning~\citep{Lazaridou2016, Havrylov2017, Kottur2017, Evtimova2017, Lazaridou2018, Bouchacourt2018, Choi2018, Korbak2019, Cogswell2019, Chaabouni2019a, Chaabouni2019b, Li&Bowling2019, Resnick2019, Guo2019, Lowe2019, Baroni2019, Chaabouni2020, Baroni2020}, as the ability to ground into other modalities a natural-like language is thought to be a prerequisite for general AI~\citep{Winograd1972,Mikolov2015,Baroni2017,Chevalier-Boisvert2018,Baroni2020}. \\


\begin{figure}[t]
    \centering
    \begin{subfigure}[t]{0.49\linewidth}
        \includegraphics[width=1.0\linewidth]{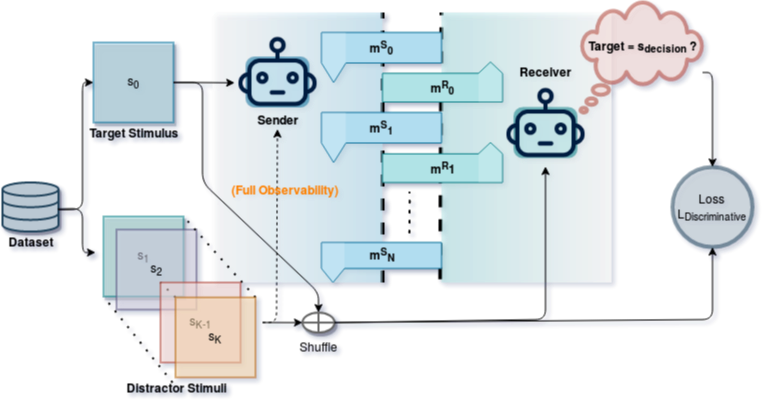}
    \end{subfigure}
    \begin{subfigure}[t]{0.49\linewidth}
        \vspace{-100pt}
        \includegraphics[width=1.0\linewidth]{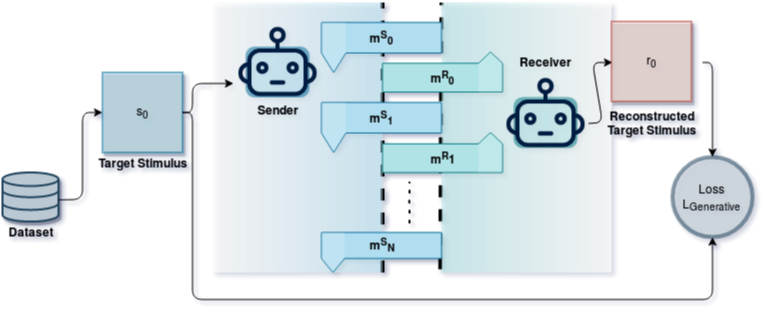}
    \end{subfigure}
    \caption{\textbf{Left:} Illustration of a partially-observable $2$-players / $K+1$-stimuli / $L$-signal / $N$-round / uniformly-distributed-distractors / stimulus-focused variant of a referential game. \textbf{Right:} Illustration of a \textit{generative $2$-players / $L$-signal / $N$-round} variant of a \textit{referential game}.}
    \label{fig:GenerativeReferentialGame}
    \label{fig:ReferentialGame}
\end{figure}

In this paper, our first contribution is to propose a nomenclature to make sense of all the surveyed initiatives, which in spite of their great variations in assumptions and constraints, can be understood under the umbrella of \textit{referential games}~\citep{lewis1969convention}. The nomenclature is presented in Section~\ref{sec:nomenclature}. Secondly, we introduce a deep learning framework entitled ReferentialGym, which is dedicated to further the exploration of language emergence and grounding in the visual modality, and their apparent properties and impact on learned representations, while providing a common implementation for all the features and constraints highlighted in the nomenclature. It is based on PyTorch \cite{pytorch-paszke-NEURIPS2019_9015} and is available under the MIT license\footnote{ReferentialGym can be downloaded at: \url{https://github.com/Near32/ReferentialGym}}. \\

The main features of ReferentialGym are:
\begin{enumerate}
    \item{ 
    It provides a \textit{common implementation for many features and constraints of the main referential game variants} that can be found in the surveyed literature to date, thus allowing for a more systematic comparison between approaches.
    }
    \item{ 
    Optimisation of the communication-channel via either \textit{REINFORCE-like algorithms}~\citep{williams1992simple}, \textit{(Straight-Through) Gumbel-Softmax estimator}~\citep{maddison2017concrete,jang2017categorical} following the work of~\citet{Havrylov2017}, or baseline and experimental variants of the \textit{Obverter approach}~\citep{Choi2018,Bogin2018,Korbak2019}.
    }
    \item{ 
    It provides implementations of the main metrics in language emergence: \textit{topographic similarity}~\citep{Brighton&Kirby2006} to evaluate language compositionality, \textit{ambiguity} to evaluate language productivity, \textit{instantaneous coordination}~\citep{jaques2018social} to evaluate communication efficiency between the agents. Also, the main metrics in representation learning are provided with a primary focus on \textit{disentanglement metrics}(e.g. ~\citep{Kim2018}), in order to investigate the impact of language emergence and grounding onto the learned representations of the visual modality.
    }
    \item{ 
    Modular graph-oriented design of the referential games allows the user to implement their own dedicated modules, for further inferences or metrics, and easily interface them with the base referential game graph.
    }
    \item{ 
    It provides out-of-the-box data management strategies for classification datasets to be experimented with in a referential game. The current framework has implemented strategies to accommodate the design principles implemented by PyTorch with regards to data management and loading, thus doing a lot of the initial heavy-lifting for the user.
    }
    \item{ 
    It provides a multi-task and transfer-learning-oriented framework where language emergence and grounding (via user-defined referential game variants) can be leveraged towards visual representation learning for other user-defined classification/regression tasks. Online training with multiple task losses are weighted via a \textit{(homoscedastic) uncertainty learning module}~\citep{Kendall2017}.
    }
\end{enumerate}

%% file: parts/referential-games.tex
\section{Referential Games Nomenclature}
\label{sec:nomenclature}

In this section, a nomenclature is proposed to make sense of the main referential game variants that have spawned in the deep learning literature recently, and have explored the problem of efficiently communicating a meaning over a (limited) communication channel, without a prior grounding of the symbols used in the communication channel. \\

The first instance of an environment that demonstrated a primary focus on the objective of communicating efficiently is the \textit{signaling game} or \textit{referential game} by ~\citet{lewis1969convention}, where a \textit{sender/speaker} is asked to send a \textit{signal/message} to the \textit{receiver/listener}, based on the \textit{state/stimulus} of the world that it observed. The \textit{receiver/listener} then acts upon the observation of the \textit{signal/message} by choosing one of the \textit{actions} available to it. Both players goals are aligned (it features \textit{pure coordination/common interests}), with the aim of performing the `best' \textit{action} given the observed \textit{state}, where the notion of `best' \textit{action} is defined by the goal/interests common to both players. \\

Variants of this game have been driving a lot of research on language emergence and communication-based co-operation in the field of  linguistics~\citep{briscoe2002linguistic, cangelosi2012simulating, spike2017minimal, steels2012grounded}, game theory\citep{crawford1982strategic, farrell1996cheap, blume1998experimental, crawford1998survey} (as acknowledged in \citep{Lazaridou2016}), and more recently, deep learning~\citep{Lazaridou2016, Havrylov2017, Kottur2017, Evtimova2017, Lazaridou2018, Bouchacourt2018, Choi2018, Korbak2019, Cogswell2019, Chaabouni2019a, Chaabouni2019b, Li&Bowling2019, Guo2019, Chaabouni2020, Ren2020}. We will focus specifically on those variants that fit in the denomination of \textit{referential games}. An instance of the kind of referential game that we are concerned with can be found in the work of \citet{Lazaridou2016}. Under the nomenclature presented in this paper, ~\citet{Lazaridou2016} would be featuring a \textit{discriminative fully-observable / $2$-players / $1$-signal / $0$-round / uniformly-distributed-distractors / stimulus-focused} variant, which they described as:

\begin{enumerate}
    \item ``There is a set of images represented by vectors ${i_1, ... , i_N}$, two images are drawn at random from this set, call them $(i_L, i_R)$, one of them is chosen to be the `target' $t \in {L,R}$
    \item There are two players, a \textit{sender} and a \textit{receiver}, each seeing the images [--thus the adjective \textit{fully-observable}, as opposed to when the \textit{sender} would only see the `target' stimulus--]- the \textit{sender} receives input $\theta_S(i_L, i_R, t)$.
    \item There is a vocabulary $V$ of size [$|V|$] and the \textit{sender} chooses one symbol to send to the receiver [--thus the adjective \textit{$1$-signalled}--], we call this the \textit{sender}’s policy $s(\theta_S(i_L, i_R, t)) \in V$.
    \item The \textit{receiver} does not know the target, but sees the \textit{sender}’s symbol and tries to guess the target image. We call this the \textit{receiver}’s policy $r(i_L, i_R, s(\theta_S(i_L, i_R, t))) \in {L,R}$.
    \item If $r(i_L, i_R, s(\theta_S(i_L, i_R, t)) = t$, that is, if the \textit{receiver} guesses the target, both players receive a payoff of 1 (win), otherwise they receive a payoff of 0 (lose).''
\end{enumerate}

\citet{Lazaridou2016} shows, firstly, that fairly simple convolutional neural networks (CNNs) can learn to coordinate via a \textit{$1$-signal} communication protocol that is learned from scratch. Secondly, \citet{Lazaridou2016} shows that ``the meanings agents come to assign to symbols in this setup capture general conceptual properties of the objects depicted in the image, rather than low-level visual properties". In other words, the relationship between the meaning/stimulus space (i.e. the pixel space) and the signal/message space (similar in shape, here, to a finite set of integers of size $|V|$) is relying strongly on the fact that CNNs are known to ``capture high-level visual properties of objects"(\citep{Lazaridou2016}, citing \citep{zeiler2014visualizing}). \\

Another variant can be seen in the work of \citet{Havrylov2017}, which would be tackling a \textit{discriminative partially-observable / $2$-players / $L$-signal / $0$-round / uniformly-distributed-distractors / stimulus-focused} variant, where the main results are presented with a number of distractors, $K=127$, and a maximum sentence length, $L=14$, and is described by ~\citet{Havrylov2017} as:

\begin{enumerate}
    \item ``There is a collection of images $\{i_n\}^N_{n=1}$ from which a target image $t$ is sampled as well as $K$ distracting images $\{d_k\}^K_{k=1}$.
    \item There are two agents: a \textit{sender} $S_{\phi}$ and a \textit{receiver} $R_{\theta}$. 
    \item After seeing the target image $t$, the \textit{sender} has to come up with a message $m_t$, which is represented by a sequence of symbols from the vocabulary $V$ of a size $|V|$. The maximum possible length of a sequence is $L$.
    \item Given the message $m_t$ and a set of images, which consists of distracting images and the target image, the goal of the \textit{receiver} is to identify the target image correctly.''
\end{enumerate}

Figure~\ref{fig:ReferentialGame} illustrates this main variant, which introduced the possibility of using a variable-length communication protocol, in addition to the reliance on the (Straight-Through) Gumbel-Softmax estimator approach (as opposed to the more common approach based on REINFORCE-like algorithms~\citep{williams1992simple}). \\ 

Following this primer, outlining the kind of language game this paper focuses on, the following sections present our tentative nomenclature by describing the main features or dimensions of variations that have been investigated to date in the literature, in order to extend the most basic \textit{referential game}.

\subsection{Full vs. Partial Observability} 
\label{sec:nomenclature:observability}

This feature characterises whether the stimuli that are presented to the \textit{sender/speaker} consist of all the stimuli experienced by the \textit{receiver/listener} or solely of the target stimulus. The basic \textit{referential game} has \textit{full observability}, and thus it allows the \textit{sender/speaker} to reason pragmatically, for instance. The orange arrow on Figure~\ref{fig:ReferentialGame} highlights additional information available when the \textit{sender/speaker} has full observability. 

\subsection{Multi-Players}
\label{sec:nomenclature:multi-players}

The basic \textit{referential game} consists of only $2$ players, one \textit{sender/speaker} and one \textit{receiver/listener}. Yet, evolutionary and computational linguistics have shown that cultural transmission (i.e. from one generation of speaker/listener agents to another) plays a major role in the emergence of properties, such as compositionality~\citep{Kirby&Hurford2002,Smith2003,Kirby2014} or recurrent syntax~\citep{Kirby199812,Kirby2015}. The work of ~\citet{Cogswell2019}, ~\citet{Guo2019}, and, most recently, ~\citet{Ren2020} illustrate implementations of implicit and/or explicit cultural transmission, where there exists at least one player/agent in each role and they may also be replaced (reset) according to different strategies. It is shown that even on the deep learning substrate, cultural transmission enhances the compositionality of the emerging language, as measured by a high performance accuracy in a referential game with novel combinations of known stimulus components, and an increase of topographic similarity between the meaning/stimulus space and the signal/message space throughout training. 
    
\subsection{Variable-length Communication} 
\label{sec:nomenclature:variable-length-communicaton}

This feature characterises the ability from the \textit{sender/speaker} to send/utter more than one \textit{symbol/signal} to the \textit{receiver/listener}, up to a maximal possible length, $L$, for the sequence of symbols. The basic \textit{referential game} is \textit{$1$-signalled}. It is first introduced by ~\citet{Havrylov2017} and is quickly adopted by the research community as standard, independently of what approach is favoured to support the communication channel ~\citep{Lazaridou2018,Choi2018}.

\subsection{Multi-Round Communication} 
\label{sec:nomenclature:multi-round}

This feature characterises whether the \textit{receiver/listener} can send a \textit{signal/message} back to the \textit{sender/speaker} and how many communication rounds can be expected before the \textit{receiver/listener} is finally tasked to act (this is described further in Section \ref{sec:nomenclature:form}). The basic \textit{referential game} is \textit{$0$-rounded}, which means the \textit{receiver/listener} is not allowed to send \textit{any} \textit{signal/message}, which could probably be understood as queries, back to the \textit{sender/speaker}. It only observes one \textit{signal/message} and thus disambiguation is not possible, and the \textit{receiver/listener} must make a correct decision on this one and only \textit{signal/message}.


\subsection{Multi-Modality}
\label{sec:nomenclature:multi-modality}

The basic \textit{referential game} is not multi-modal in the sense that both the \textit{sender/speaker} and \textit{receiver/listener} experience stimuli through the same modality (usually either raw pixel inputs or symbolic (one-hot encoded) inputs). On the other hand, \citet{Evtimova2017} featured different modalities, one for each role, such that the \textit{receiver/listener} (referred to as the \textit{questioner}) is experiencing raw pixel inputs while the \textit{sender/speaker} (referred to as the \textit{answerer}) is experiencing textual descriptions (in natural language). This work instantiates a \textit{discriminative multi-modal / partially-observable / $2$-player / $L$-signal / $R$-round / all-distractors / stimulus-focused} variant of a \textit{referential game}. \\

\subsection{Discriminative vs. Generative}
\label{sec:nomenclature:form}

As mentioned earlier in Section \ref{sec:nomenclature:multi-round}, after all the communication rounds have been performed, the \textit{receiver/listener} is tasked with performing an action. Depending on the type of referential game, the action can take two main forms:

\begin{itemize}
    \item{ 
    \textbf{Discriminative} - In this form, the agent has to discriminate between a set of stimuli, comprised of the target stimulus observed by the \textit{sender/speaker} and some additional distractor stimuli, and find the target.
    }
    \item{ 
    \textbf{Generative} - In this form, the agent has to generate an output, which can for instance, be the task of reconstructing the target stimulus itself or some of its (symbolic) attributes~\citep{Kottur2017,Chaabouni2020}.
    }
\end{itemize}

While the generative form is rather simplistic, in terms of the number of moving parts, as illustrated in Figure~\ref{fig:GenerativeReferentialGame}, it is not the case for the discriminative form. The following subsections describe the different refinements that can be found for the discriminative form in the literature. 

\subsubsection{Distractor Stimuli Distribution} 
\label{sec:nomenclature:form:distractor-distribution}

This feature characterises the kind of distribution from which the distractor stimuli are sampled and presented to the \textit{receiver/listener}. The basic \textit{referential game} has distractors \textit{uniformly-distributed}. ~\citet{Lazaridou2018} investigated the effect of sampling distractors from a ``target-specific context distribution'' that emphasizes how common objects actually co-occur in the real world, thus making ``the target \textit{goat} more likely being mixed with \textit{sheep} and \textit{cow} as distractors rather than \textit{bike} or \textit{eggplant}''. They reported that, in this non-uniform case, the game is made significantly more difficult, reporting lower data-efficiency, as similar stimuli co-occur more often. More importantly, this feature influenced the ``organisation''/structure of the emerging languages. \citet{Lazaridou2018} opens an avenue to explore the naturalness of emerging languages, as it is a growing concern in the community~\citep{Kottur2017,Chaabouni2019a,Chaabouni2019b,Baroni2020}, in view of better human-machine interface.

\subsubsection{Descriptive-only} 
\label{sec:nomenclature:form:descriptive}

The basic (discriminative) \textit{referential game} allows the \textit{receiver/listener} to perform pragmatic reasoning on its stimuli, since it experiences at least two stimuli, the target stimulus and at least one distractor stimulus. In this variant, the \textit{receiver/listener} only experiences one stimulus that \textit{may or may not be} the target stimulus, and it is tasked to output whether the stimulus it has experienced is the same as the \textit{sender/speaker} has. Only the descriptive-part of the \textit{referential game} is emphasised here. It is argued that the work of \citet{Choi2018} deals with such a variant, described as the \textit{Two-Person Image Description Game}. The effect of this setting has not yet been compared to other \textit{referential game} variants.

\subsubsection{Stimulus vs. Object Focused} 
\label{sec:nomenclature:form:focus}

The basic (discriminative) \textit{referential game} is stimulus-focused, which assumes that both agents would be somehow embodied in the same body, and they are tasked to discriminate between given stimuli. On the other hand, the object-focused variant incorporates the issues that stem from the difference of embodiment. The agents are tasked with discriminating between objects (or scenes) independently of the viewpoint from which they may experience it. In this variant, the game is more about bridging the gap between each other's cognition rather than (just) finding a common language. It is solely featured in the work of \citet{Choi2018}, in its descriptive-only form. Needless to say that the object-focused variant adds a considerable degree of difficulty to the task. It has been highlighted that embodiment may hold some key to the systematic generalisation abilities of the learning agent~\citep{Hill2019}, and therefore it is highlighted as a very important research direction to pursue.\\

\citet{Choi2018} shows that the obverter technique enforces great concept alignment between the two agents (to which extent do the (visual) sensory features align from one agent to another agent, when projected into a similar (linguistic) space?), here aligning the languages spoken by both agents and also aligning the way modalities are internally represented. Interesting research directions to highlight dwell in (i) the evaluation of the efficiency of the other agents architecture in this object-focused setting, and (ii) the agent-to-agent concept/feature/modality representation alignment. \\

In an even more abstract approach, the object-focused setting could be acknowledged as an emphasis on the concept or semantic meaning behind the observed stimulus, and the \textit{receiver/listener} would thus be tasked with learning the semantic, while being prompted with different instances of it. As detailed further in Section \ref{sec:referential-gym}, this viewpoint opens up numerous applications of referential games for classification tasks. \\

\subsection{Limitations}
\label{sec:nomenclature:limitations}

Finally, with regards to some other important examples, it can be argued that the works of \citet{Das2017,Kottur2017,Cogswell2019} also feature \textit{generative multi-modal / partially-observable / $2/N$-players / $1/L$-signal / $R$-round} variants, referred to as the \textit{Task \& Talk Game} or as a \textit{Goal-Driven Neural Dialog}. Indeed, the modality of the \textit{receiver/listener} (\textit{questioner}) is different from that of the \textit{sender/speaker} (\textit{answerer)}, since the latter experiences a one-hot-encoded vector describing the task to solve, i.e. to find the answer to an attribute-focused question about the stimulus experienced by the \textit{sender/speaker}. For example, the question can concern the value instantiated in the \textit{shape}, \textit{color}, or \textit{style} attribute of said stimulus, that can represent for instance a \textit{dashed green circle}. 
It is important to note that although these games are not explicitly implemented as referential games, they could be instantiated as a generative referential game variant. \\

Furthermore, although an explicit difference is made between the generative and discriminative forms, it can be noticed that the generative form is implicitly instantiating a discriminative form, where the \textit{receiver/listener} modality consists of the space of all possible stimuli. For instance, in the case of the \textit{Task \& Talk Game}, this space consists of all the possible values permitted on each attribute axis. It ensues that the latter agent would be tasked to choose the value of the queried attribute that is instantiated in the stimulus observed by the \textit{sender/speaker}, among the whole collection of values that can be instantiated for each queryable attribute of the stimulus.
In the nomenclature presented in this paper, the decision was made to make an explicit difference between the two forms. Each form entails a different practical framing of the task at the level of the agent, which is known to have an impact on the agent's performance at any given task. This has given rise to sub-fields in deep learning literature, e.g. the discriminative form is close to adversarial deep learning, whilst the generative form is close to auto-regressive/generative deep learning.

%% file: parts/referential-gym.tex
\section{ReferentialGym Architecture}
\label{sec:referential-gym}

Based on the previously-detailed nomenclature in Section \ref{sec:nomenclature}, ReferentialGym was developed to provide a common framework for improving comparisons between architectures when investigating the impact of features on emerging languages and their properties. ReferentialGym aims at providing a coherent framework where such comparisons can be undertaken quickly, easily and fairly, as much of the heavy-lifting has already been done, so users can focus on prototyping and experimenting with different architectures. In the remainder of this section, a walkthrough is presented of the architecture and design principles at the heart of the ReferentialGym framework.\\

The features identified by the nomenclature are associated to some hyperparameter entrypoints that the user is asked to set when defining the kind of referential game to instantiate. 
It takes the form of an instance of the class {\tt ReferentialGame} that handles the training and testing phases, in multiple epochs (with a different time granularity within each epoch), on a data set of the user's choosing. The only limitation is that this data set ought to be provided as an instance of the PyTorch {\tt torch.utils.data.dataset} class. It is then internally wrapped multiple times to accommodate sampling the stimuli of the target and distractors, which depends on the form of the game the user has defined, among the many possibilities highlighted in the nomenclature defined in Section \ref{sec:nomenclature}, when setting the hyperparameters of the {\tt ReferentialGame} instance. \\

It is important to note that the different internal wrappers are handling all the data management strategies, in order to allow any image classification data set to be used for experiments in most of the referential game variants. Notably, different distractor sampling strategies are made available with respect to the class of each stimulus in the classification data set. Object-focus can also be enforced in the form of a focus on the semantic or the class of each image, as images of the same class can be seen as a different viewpoint or instance of the same semantic scene/object (as we previously highlighted in Section~\ref{sec:nomenclature:form:focus}). 

\begin{figure}[t]
    \centering
    \includegraphics[width=0.65\linewidth]{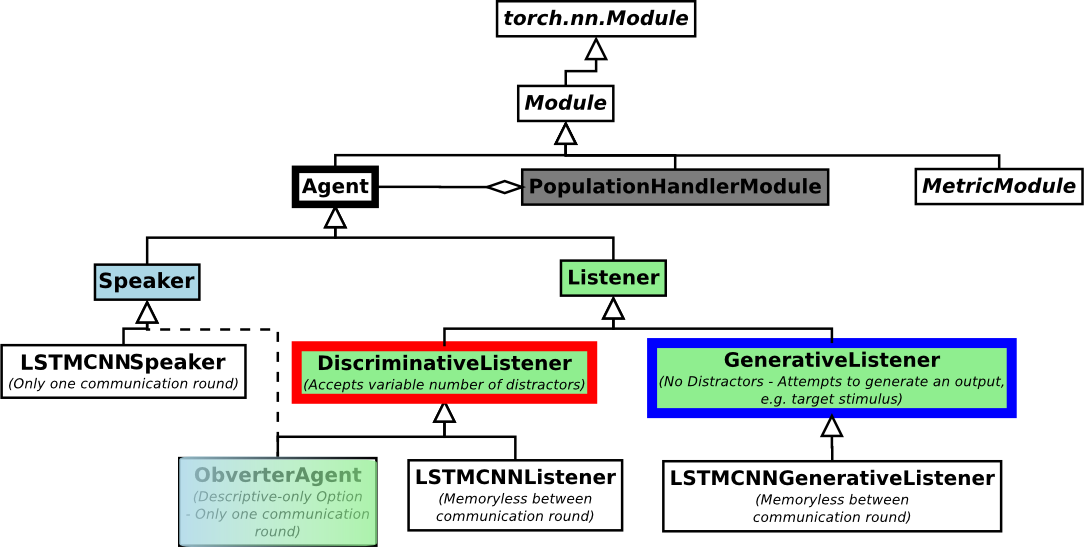}
    \caption{Hierarchy of referential game agents implemented in ReferentialGym and their relationship to the more general {\tt Module} class.}
    \label{fig:referentialGym-agent-hierarchy}
\end{figure}

\subsection{Data Streaming-Oriented Design, Logging \& Metrics}
\label{sec:referential-gym:design}

The framework is built with modularity and re-usability in mind, therefore it relies on modular building blocks interconnected in acyclic graphs, where each building blocks inherits from a {\tt Module} class (itself inherited from PyTorch's {\tt torch.nn.Module} class) that requires the definition of input data streams on which to operate. The output data streams produced by each {\tt Module} instance are then handled by an instance of the {\tt StreamHandler} class that is in charge of serving each module with their desired input data streams. \\

To the user, all of this is hidden and happens seemingly in the background, similarly to some statistics logging and agent management processes (e.g. saving, loading, etc). The whole framework implements logging mechanisms  relying on TensorboardX~\citep{huang2018tensorboardx}, which are then visualizable in Tensorboard~\citep{manetensorboard}. \\

In addition to statistics logging, the framework provides implementations of major metrics in the field of language emergence, accounting for both \textit{positive signalling} and \textit{positive listening} metrics, following the work of ~\citet{Lowe2019}. These include:  \textit{topographic similarity}~\citep{Brighton&Kirby2006}, in the form of the {\tt TopographicSimilarityMetricModule} class; \textit{ambiguity} in the emerging language, in the form of a hook function attached to each {\tt Speaker} module (see Section~\ref{sec:referential-gym:agents} for more details about hooking mechanisms); and \textit{instantaneous coordination}~\citep{jaques2018social}, in the form of the {\tt InstantaneousCoordinationMetric\\Module} class. Moreover, as the framework lies at the interface between language emergence and language grounding in the visual modality, it aims to provide the means to evaluate the learned representations at different levels, not only the level of language, but also the level of the visual modality. In this initial release, this is achieved by providing implementations of \textit{disentanglement metrics}, in the form of the {\tt FactorVAEDisentanglementMetric\\Module} that follows the metric proposed by ~\citet{Kim2018}. 

Finally, each acyclic graph is defined as a pipeline, or ordered list of modules that the {\tt StreamHandler} instance serves at each timestep. Pipelines are provided by the user as ordered list of module identifications.

\subsection{Agent Hierarchy}
\label{sec:referential-gym:agents}

Everything in the framework inherits from the {\tt Module} class. This is evidently the case of referential game agents whose inheritance hierarchy is illustrated in Figure~\ref{fig:referentialGym-agent-hierarchy}. Following the detailed nomenclature in Section \ref{sec:nomenclature}, while the \textit{sender/speaker} is invariant to the form that the game takes (and thus is always instantiated in the {\tt Speaker} class), the \textit{receiver/listener} computation obviously varies. Playing in the discriminative form requires the instantiation of the \textit{receiver/listener} in the {\tt DiscriminativeListener} class hierarchy, whereas playing in the generative form requires its instantiation to be in the {\tt GenerativeListener} class hierarchy.

In more detail, classes that inherit from the {\tt Agent} class come with a hooking mechanism that enables the user to augment the functionality of each pre-defined agent, by defining and registering hook functions that would operate on the output data stream of the agents, before the {\tt StreamHandler} sees them. This hooking mechanism is primarily used in the background to define different loss functions that accommodate the different referential game variants, and different logging mechanisms that may, for instance, be specific to the agent type (e.g. ambiguity metric specific to {\tt Speaker} agents, as mentioned earlier). \\ 

The framework also implements a very powerful abstraction, in which modules can also be considered as possible input data streams, thereby allowing some modules to operate on some other modules. The main purpose of this abstraction is to enable multi-player variants of the game, in addition to different management approaches to the culture of agents if the user chooses so, via an instance of the {\tt PopulationHandlerModule} class, whose output streams are placeholder agents that are subsequently playing a round of the referential game.



%% file: parts/related-future-works.tex
\section{Related \& Future Works}

\textbf{Language Grounding} - 
The proposed framework is inspired by the possibilities offered by the PyTorchPipe (PTP) framework~\citep{kornuta2019pytorchpipe} and aims to provide a similar tool towards investigating artificial language emergence and, in latter releases, investigating translation between natural and artificial languages. ReferentialGym and PTP both operate in the language grounding subfield, among others, and share some design principles, as they are both focusing on the modularity, and therefore re-usability of each component, which can be arranged by the user. \\

\textbf{Language Emergence} - 
In the current release, ReferentialGym is grounded in a broader range of features found in the relevant literature than the EGG framework~\cite{kharitonov2019egg}, albeit substantially less mature. ReferentialGym is mainly focused with visual stimuli, while the EGG framework remains rather general on that side. Going forward, ReferentialGym will focus on dynamic stimuli, in the form of videos, and plans to accommodate video data sets in order to investigate language emergence and grounding over transformations or time-sensitive data/stimuli. \\ 

\textbf{Disentanglement \& Compositionality} - In the current release, ReferentialGym provides modular implementations of disentanglement and compositionality metrics, in addition to state-of-the-art autoregressive inference modules, such as: $\beta$-VAE and variants~\citep{Higgins2016,Burgess2018}, FactorVAE~\citep{Kim2018}, and MONet~\citep{burgess2019monet}. Integration of relevant datasets to study the relationship between disentanglement in the learned representations and compositionality in the emerging languages are well underway, starting with the dSprites~\citep{Higgins2016,dsprites17} data set.

%% file: parts/conclusion.tex
\section{Conclusion}

This paper provides two main contributions to the research community. Firstly, a nomenclature is proposed for the different language games that have spawned under the umbrella of referential games with the goal of studying language emergence and grounding. The many features and varying constraints that the literature exhibits have been discussed, and their relevance towards the development of language emergence and grounding abilities, as prerequisites for general AI, have been highlighted. 
Secondly, this paper introduces ReferentialGym, a deep learning framework that implements the main features of the proposed nomenclature. ReferentialGym is based on PyTorch\cite{pytorch-paszke-NEURIPS2019_9015} and dedicated to further the exploration of language emergence and grounding in the visual modality, by providing baseline implementations of major algorithms and metrics of the surveyed literature. It is hoped that this work will ease the entry barrier to the field and enable the community to perform more thorough and fair comparisons, thanks to the collection of common implementations.